# Autonomous Systems: Indoor Drone Navigation

Authors
Aswin Iyer, Santosh Narayan, Naren M, Manoj kumar Rajagopal

Institution
Vellore Institute Of Technology, Chennai Campus

First Author, Second Author, Third and Fourth Author:
iyerashwin.raja2019@vitstudent.ac.in, naren.m2019@vitstudent.ac.in,
santoshnarayan.v2019@vitstudent.ac.in, manojkumar.r@vit.ac.in

## ABSTRACT:

DRONES ARE A PROMISING TECHNOLOGY FOR AUTONOMOUS DATA COLLECTION AND INDOOR SENSING. IN SITUATIONS WHEN HUMAN-CONTROLLED UAVS MAY NOT BE PRACTICAL OR DEPENDABLE, SUCH AS IN UNCHARTED OR DANGEROUS LOCATIONS, THE USAGE OF AUTONOMOUS UAVS OFFERS FLEXIBILITY, COST SAVINGS, AND REDUCED RISK. THE SYSTEM CREATES A SIMULATED QUADCOPTER CAPABLE OF AUTONOMOUSLY TRAVELLING IN AN INDOOR ENVIRONMENT USING THE GAZEBO SIMULATION TOOL AND THE ROS NAVIGATION SYSTEM FRAMEWORK KNOWN AS NAVIGAATION2. WHILE NAV2 HAS SUCCESSFULLY SHOWN THE FUNCTIONING OF AUTONOMOUS NAVIGATION IN TERRESTRIAL ROBOTS AND VEHICLES, THE SAME HASN'T BEEN ACCOMPLISHED WITH UNMANNED AERIAL VEHICLES AND STILL HAS TO BE DONE. THE GOAL IS TO USE THE SLAM TOOLBOX FOR ROS AND THE NAV2 NAVIGATION SYSTEM FRAMEWORK TO CONSTRUCT A SIMULATED DRONE THAT CAN MOVE AUTONOMOUSLY IN AN INDOOR (GPS-LESS) ENVIRONMENT.

## SECTION I: INTRODUCTION:

In recent years, there has been a growing demand for autonomous systems in various domains, including robotics. Autonomous systems in constrained environments, such as indoor environments, have gained immense popularity due to their potential value in applications such as emergency rescue operations, inspection, inventory management, and surveillance in industrial laboratories, to name a few. Unmanned Aerial Vehicles (UAVs), commonly known as drones, have emerged as a promising solution for autonomous data gathering and sensing in indoor environments without human intervention. The use of autonomous UAVs offers flexibility, cost savings, and reduced risk in scenarios where human-controlled UAVs may not be feasible or reliable, such as in unknown or hazardous environments.

One of the critical challenges in enabling autonomous navigation of UAVs in indoor environments is the ability to accurately perceive and map the surroundings. Mapping and localization are crucial processes that provide the UAV with information about the environment and its position relative to it. Mapping involves gathering information about nearby obstacles and available pathways to traverse, while localization determines the UAV's current location with respect to a defined frame of reference. Sensors such as cameras, ultrasonic sensors, and laser scanners are commonly used to measure the distance between the UAV and obstacles.

In this research paper, we present a comprehensive approach for the autonomous navigation of a UAV in an indoor environment in a Robotic Operating System (ROS). The system utilizes Nav2, a navigation system framework available in ROS, and the Gazebo simulation tool, to develop a simulated quadcopter UAV capable of autonomously navigating in an indoor environment. LiDAR technology is used for scanning the room, and a map of the environment is created with the help of SLAM (Simultaneous Localization and Mapping), using SLAM Toolbox for ROS. The map is then used for the quadcopter's autonomous navigation, allowing it to avoid obstacles and travel from one location to another safely and efficiently. While the successful operation of autonomous navigation in ground vehicles and land-based robots has been demonstrated using Nav2, the same hasn't been achieved with UAVs and has a long way to go. The objective of this research is to build a simulated drone, which can travel autonomously, in an indoor (GPS-less) scenario, using the Nav2 navigation system framework and SLAM Toolbox for ROS. It can also be thought of as an interesting use case for demonstrating the extension of functionalities of the existing navigation system Nav2 for ground robots to UAVs.

## SECTION 2: RELATED WORK:

[1] describes a robot navigation implementation of a Parrot AR and a robotic operating system based on autonomous simultaneous localization and mapping (SLAM). Parrot AR Drone 2.0 is equipped with an inertial measurement unit and a laser scanner. The global planner employs the A* search algorithm, whereas the local planner makes use of the dynamic window strategy. The outcomes of the Gazebo simulation confirm the capable performance of both the suggested SLAM and the offered navigation algorithm.

[2] showed that the reinforced concrete structure of the building easily blocks indoor satellite signals, making it impossible to provide accurate real-time positioning information. This severely restricts the drone's ability to perform in special environments like indoors. So in this paper, using a LIDAR-based indoor navigation system, the issue of indoor structures obstructing satellite signals is resolved. Effective real-time positioning information is provided via synchronised positioning and mapping techniques. The drone is freed from its reliance on satellite signals, enabling it to generate its maps and navigate autonomously in unique conditions like indoors.

[3] describes an experiment to evaluate the effectiveness of drone navigation using dead reckoning. Based on IMU data, the position was calculated. Indoor drones have been demonstrated to have weight and size restrictions. The drone's ability to carry certain types of navigational sensors and processing power is impacted by this circumstance. As a result, in this case, a virtual 3D map was employed as a point of reference to ascertain the actual position of the drone. There were four alternative flight trajectories used for the experiment.

In [4] emphasizing small drones, they offer an overview of the current state of the art and unresolved issues in the field of indoor drones. Over the past few decades, indoor navigation has been a prominent area of drone study. The name "indoor" was used primarily because, unlike outside situations, indoor environments don't allow drones to rely on global navigation systems like GPS for their position and velocity calculations.

In [5] they developed an image processing algorithm and successfully interfaced RPi with Pixhawk Mini autopilot using Bluetooth telemetry. They were able to conduct a test flight that demonstrates autonomous navigation and landing of the prototype under study using Python programming and Ardupilot software and establishing short-range telemetry communication between the two using Bluetooth modules.

## SECTION 3: METHODOLOGY:

## A. Platform Used:

This research takes place in ROS, or Robot Operating System, an open-source middleware framework that provides a collection of software libraries and tools for developing robot applications. ROS provides a distributed computing environment that enables communication between different components of a robot system. One of the key advantages of ROS is its flexibility. It provides a range of libraries and tools for different robot applications, including navigation, perception, manipulation, and control. It also provides visualization and simulation tools such as Gazebo and Rviz, which will be discussed in detail further down, that allow users to test and evaluate their robots in a virtual environment before deploying them in the real world. Its modular design and message-passing infrastructure make it easy to develop and integrate new components, and its active community provides a wealth of resources and support for developers.

## B. Simulation Environment:

The quadcopter is built, operated and designed using Gazebo. Gazebo is a free and open-source 3D simulation environment that is commonly used with ROS for robotics applications. It provides a realistic simulation environment for testing and developing robots. It offers a variety of features that make it well-suited for UAV simulations. It supports physics-based simulations, multiple sensor and camera types, and the ability to simulate diverse environments and scenarios. Additionally, it has an intuitive interface that enables users to interact with their simulations in real time and visualize the outcomes. When integrated with ROS, Gazebo can work in sync with ROS nodes and messages, enabling developers to simulate and test robots. This integration permits the testing of robot control algorithms, and sensor fusion algorithms in a simulated environment before deploying them on real robots. This versatility makes it an ideal choice for testing and developing complex robotics applications such as the objective of this research.

## C. Model Specification:

The robot (UAV) model is described using its physical parameters. URDF (Unified Robot Description Format) files are created to specify the robot's dimensions, visual aspects, collision aspects and the type of movement each part undergoes such as the transmission elements of the rotors. Xacro tools are used to simplify the creation and management of URDF files and help enable modularity. Different elements of the robot are organized in separate files, which helps with debugging and preventing code clutter. For example, in the proposed model, a laser scanner (LiDAR) is attached to the bottom and kept separate from the robot's base frame description, so that it can be handled and debugged separately without interfering with the drone's base frame. An image of the drone design is shown below in Fig 1.

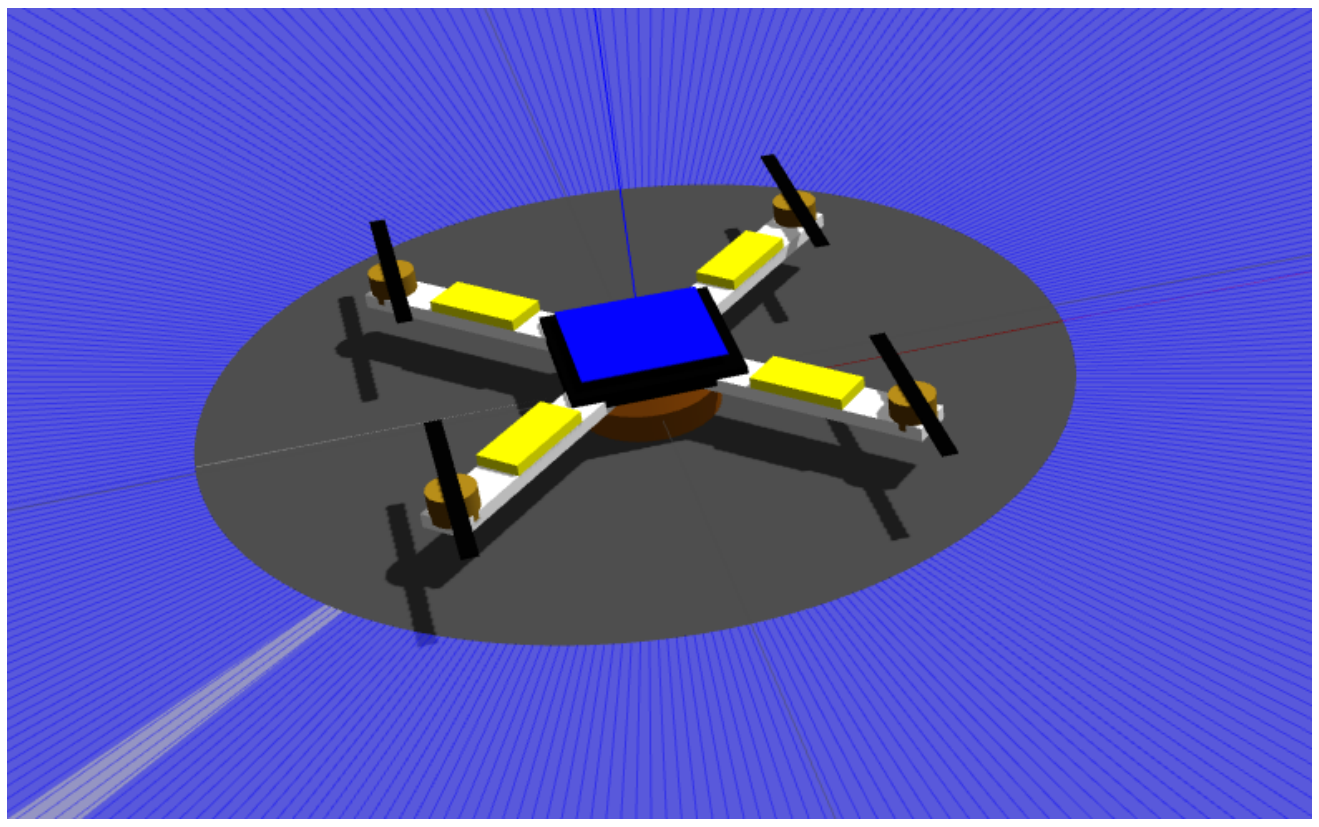

(Fig 1: Simulated Drone)

The mathematics of robotics heavily relies on coordinate transformations or transforms. They are mathematical tools for taking measurements or points that are represented from one point of view and representing them from another point of view. Without transformations, one would have to use trigonometry to complete the computations for each transform, which soon gets difficult with bigger problems, particularly in 3D. The robot state publisher package and its nodes help us visualize where our robot is in space and where all the parts of the robot are with respect to the quadcopter's base frame of reference. So whenever the motors spin to make the UAV move, the transforms to the other parts of the drone are published by this node.

The parallelograms in Fig 2 are ROS topics. These hold information that nodes can decide to publish or subscribe to it, thereby enabling communication.

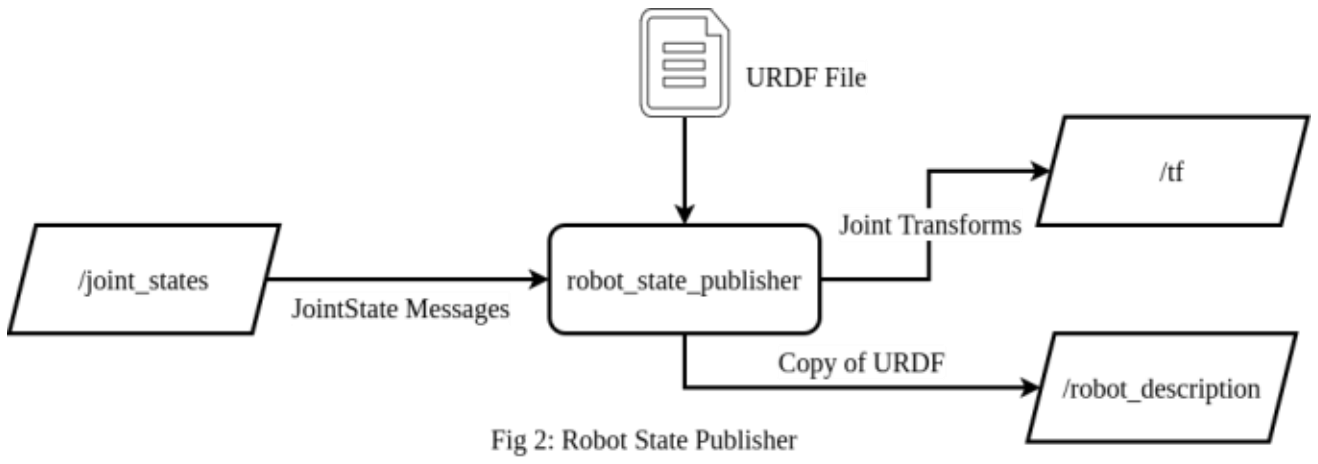

Fig 2: Robot State Publisher

(Fig 2: RSP image)

Whenever Gazebo wants to communicate with ROS (or any other external software), plugins are required. These are additional code libraries that can be installed, referenced in our URDF, and then have Gazebo run as necessary. The plugins being used can be seen in Fig 3. LiDAR plugin is used for simulating LiDAR data, ros2_control plugin for controlling the robot which will be explored in detail in later sections and a joint state publisher, which tracks all the joints in the simulation environment and feeds it to the robot state publisher as seen before. The robot spawner picks up the robot description present in the URDF file and places the robot inside the gazebo. Since all these tasks have to be done repetitively every time the simulation is launched, i.e. launching different nodes, providing parameters and input files to it, a launch script is created which takes care of all the initializations and defines all these tasks. So the simulation can directly be started with a single command with the launch file anytime without going through the same process again.

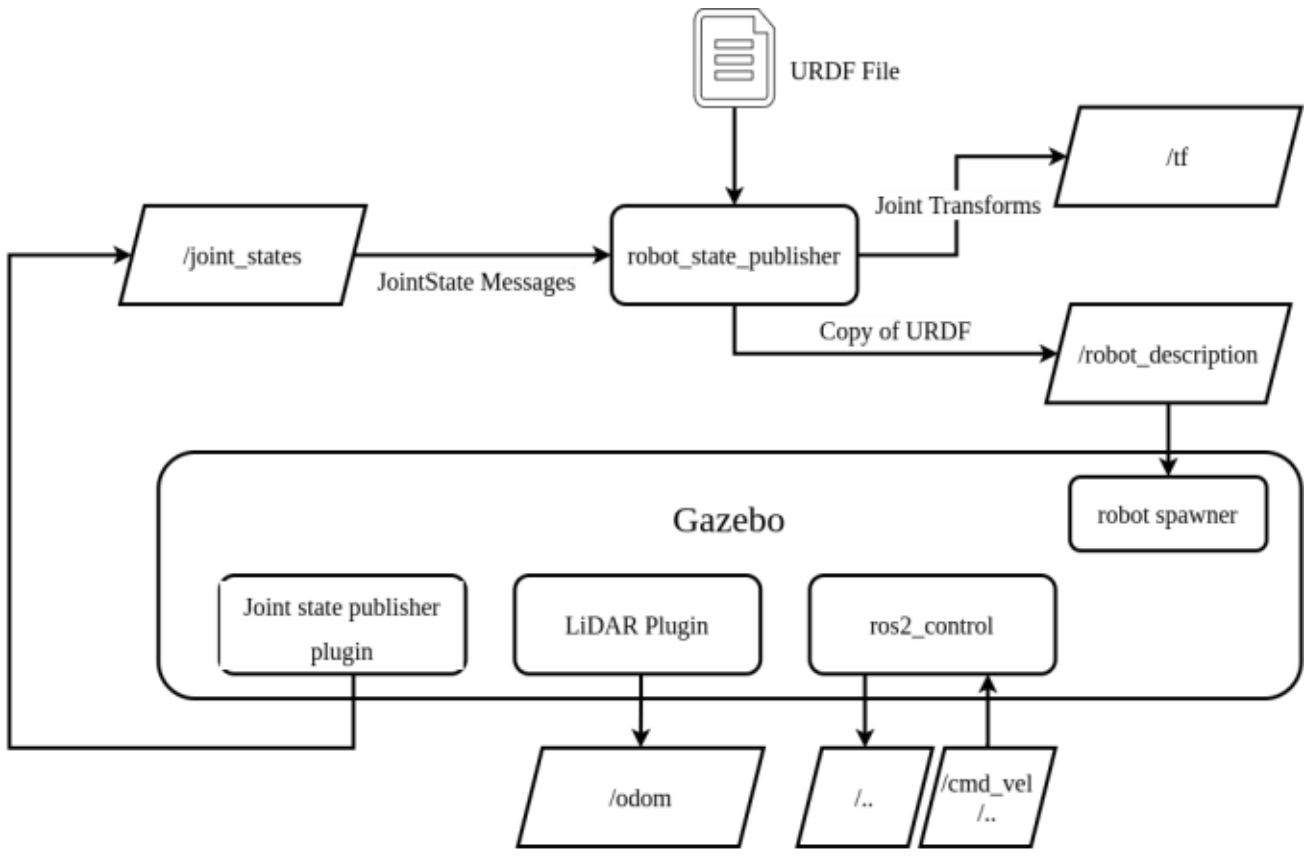

Fig 3: Gazebo-Ros Architecture

(Fig 3: Architecture)

A sample environment is created in the Gazebo for the robot to move and map accordingly. The sample map is shown in Fig 4. In this environment, several objects were placed randomly where the map is created. These objects are considered static.

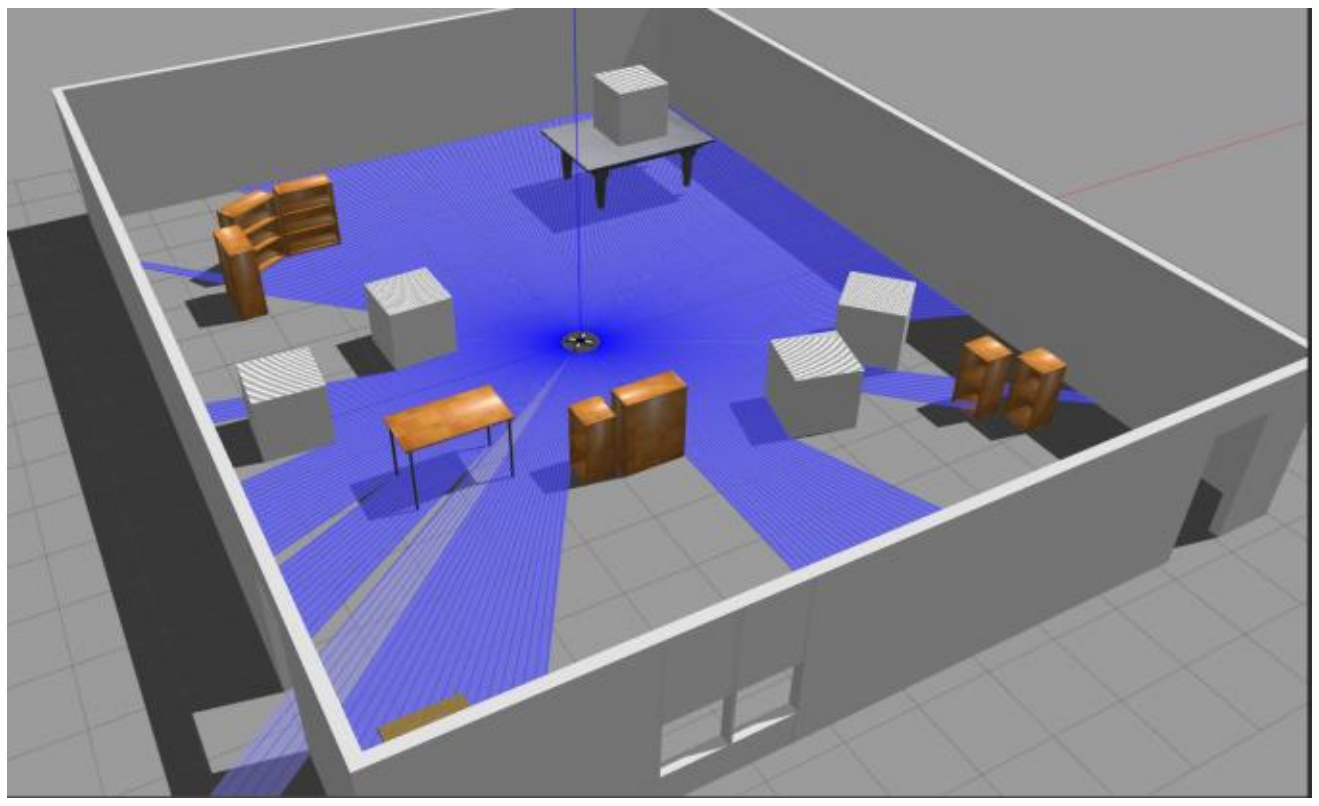

(Fig 4: Simulated Environment)

### D. Controlling the Robot:

Every robot's basic function is to receive some sort of input (from the user or the environment), process that information, and then drive an actuator, which could be a motor, a hydraulic system, or another component. There are numerous types of actuators and interfaces available, which can employ a wide range of control approaches to address issues with them. However, there is a lot of commonality within this vast array of options, so if one doesn't have some sort of standardized system and simply writes a custom controller and custom interface for every application, time is wasted rewriting things and fixing issues that people have already fixed numerous times. Instead, a framework is needed in which algorithms for various control techniques and hardware platforms are already designed, and have them speak a common language.

The ros2_control [6] package is used as it makes reusing code, upgrading, and modifying systems way easier. The controller manager locates and connects each piece of code for our hardware drivers and controllers. It uses a plugin system to accomplish this, so none of these things is running their own executables; instead, they are essentially libraries that are loaded at runtime and have a set of functions that will link into the system. Each hardware requires various controls. Whatever the hardware looks like, it needs a hardware interface to work with ros2 control. This piece of code communicates with the hardware and exposes it in accordance with the ros2 control standard. As users, all one needs to understand is how the hardware interface represents their hardware, which is through command interfaces and state interfaces. The hardware interface serves as an abstraction in this regard. command interfaces can be controlled, whereas state interfaces can only be observed. In our case motors are controlled using velocities which form the command interfaces and using actuator feedback the motor position and velocities are retrieved. The controllers only see a long list of command and state interfaces thanks to the resource manager, which the controller manager utilizes to collect and expose all of the hardware interfaces. They are added to the URDF file using a <ros2 control> tag since they are closely related to the hardware design of the robot.

The controllers are how ros2 control communicates with the rest of the ROS ecosystem. They will be listening to a ROS topic on one end for control input, such as joint positions or mobile body velocities. With this input, they will run algorithms to determine the proper actuator speeds, positions, etc., which they will then pass to the proper hardware interfaces.

Here, the controller manager's responsibility is to match the controllers that it is asked to load with the appropriate command and state interfaces that the resource manager is exposing. A YAML file containing the necessary parameters is created and sent to the controller manager to set up the controllers. The controllers can be started and stopped as necessary once everything has been loaded.

The gazebo_ros2_control plugin used here runs its own controller manager hence a standalone controller manager is not needed. Then, once the controller manager is running, interaction is possible with it to do things like checking the hardware interfaces and starting the controllers. Fig 5 gives the architecture of the ros2_control specifically for our case.

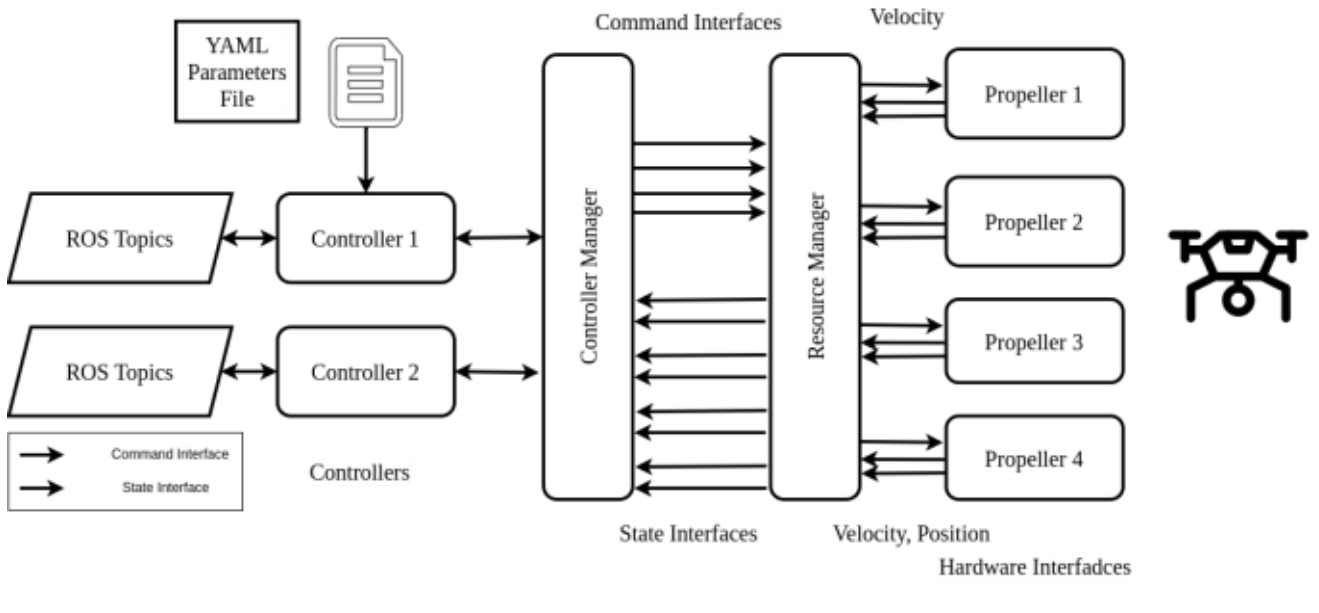

Fig 5: ros2_control

(Fig 5: ros2_control architecture)

The hardware interface is taken care of by the Joint Velocity Group Controller. Only the joint velocities are given to it and the interface will take care of actuating it on the propellers. PID control is used to indicate the combined effort by controlling the velocity. The controller first demands a velocity of zero.

Coming to control of the drone, a PID controller is used to control the standard flight parameters, which are thrust, roll, pitch and yaw. With these 4 parameters or what are called degrees of freedom propeller velocities can be calculated. For the other 2 degrees of freedom, the movement in the x and movement in the y direction is done simply by use of pitch and roll movements simply such that the component of it in the z direction remains constant to ensure thrust i.e. the upward component remains the same and the perpendicular component provides the movement.

PID (Proportional-Integral-Derivative) control is a common method that can provide stable and accurate control. It involves calculating an error signal by comparing the desired output (setpoint) with the actual output (process variable) of the drone. The error signal is then used to adjust the control inputs to bring the drone closer to the desired setpoint. The control outputs from these three terms are combined to produce the final control signal that is sent to the drone's actuators to adjust its position or orientation. The PID controller can be tuned by adjusting the gains of the P, I, and D terms to optimize the drone's response.

## E. Localization and Mapping:

The location data of ground robots are generally calculated using simple math involving the radius of the wheel and wheel separations which then combined with the rpm of the wheel one can calculate the current position of the robot from the origin. This origin is the gazebo origin or the reference point set in the world frame of reference. Since LiDAR attached to the drone is used, the origin becomes the drone, and whatever is seen in the odometry data, is from the drone's frame of reference. For obtaining odometry from the gazebo or global frame of reference, methods used on ground robots cannot be used as propeller radius and rpm are useless for this matter and cannot determine the position of the drone in the real world. Hence a different mechanism is applied to get the drone's position in the real-world frame of reference.

The rf2o laser odometry [7] package is a high-precision 3D LiDAR odometry solution that can be used for localization and mapping in drones which can solve our problem. It uses Range Flow 2D Laser Odometry algorithms and minimizes a robust function to accurately estimate the drone's position and orientation in real-time

The odometry from the global frame of reference is needed to perform SLAM. The robot or the drone doesn't need to be aware of the local environment, all this work is processed at the base station which will then direct the robot or the drone on where to go next. This will be based on the map that was generated using SLAM. It can be seen from Fig 6 that the drone has moved

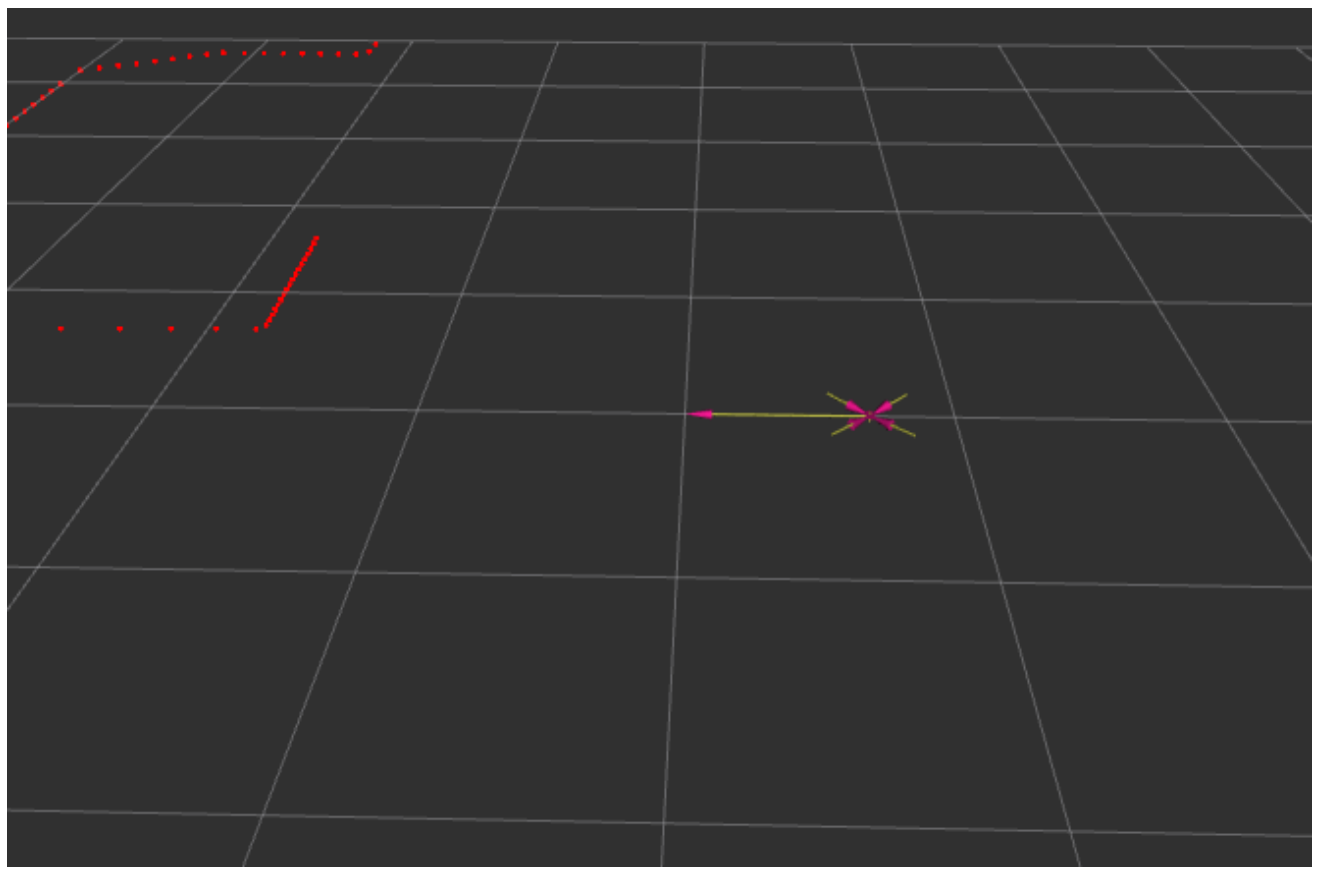

(Fig 6: Global fixed reference Odom)

SLAM is used for mapping the environment in which the robot can travel as mentioned before. The package used is the SLAM toolbox [8]. This tool can learn and create occupancy grid maps using laser scan data in a highly efficient manner. This type of map is where the existing space is divided into grids of smaller sizes and is used like a coordinate plane. Greyscale is used to denote whether the grid is occupied or not. First, a static map is created by moving the robot manually to figure out where the obstacles are present in the room.

A controller has been designed and programmed so that the flight of the drone remains stable after takeoff. It has been programmed to lift off hover at a point. This now reduces the dimensionality of the problem. Ignoring the z-axis, the drone can now manoeuver in a 2D plane at the level at which the drone is hovering. Since this is a 2D plane to deal with, all the capabilities of the SLAM package in 2D can be utilized. The slight technicality involved here is that a footprint link is added in the robot descriptions since the drone is also flying on the z-axis there has to be a reference to the absolute origin of the environment, which is essentially the footprint of the drone.

There are different modes in which the SLAM toolbox can be run, the one chosen is the Online Asynchronous mode. Online means the decisions made are based on the live data streams rather than logged data and Asynchronous mode means that our processing rate is slower than the scan rate. This is sufficient as the latest data is more important than the amount of data. So the occupancy grid map generated in the 2D plane is then used by the Navigation system that is discussed in the next section.

A Map server is created using the Nav2 package to run AMCL [9] (Adaptive Monte Carlo Localization). This capability is also provided by the Nav2 which takes in the laser scan messages and generates a transform between the occupancy map and the odometry which is required for navigation. A map to odometry transform is published which makes this Localization possible as shown in Fig 7.

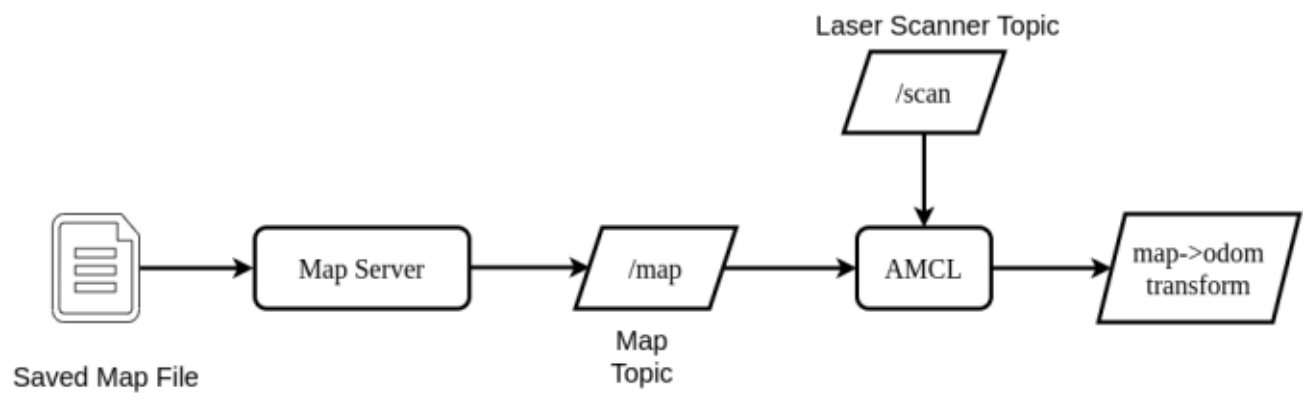

(Fig 7: AMCL image)

Then a cost map is generated, which is used for navigation, where the cost is highest wherever an obstacle is present, lowest in free spaces, and a gradient around and between the two. This implies that the cost for the robot to traverse where obstacles are present is high, and less expensive to do so in free spaces. The same map is used in the Nav2 [10] stack.

## F. Navigation:

For navigation, a safe trajectory is planned and executed from an initial pose to a target position. Nav2 (Navigation2) provides a set of tools and algorithms for autonomous navigation in robots. It is built on top of ROS and provides a range of features such as Path Planning, Obstacle, Avoidance, Behaviour Tree, etc. It provides a path planner that can generate collision-free paths avoiding obstacles for a robot to follow, taking into account obstacles and the robot's capabilities. The architecture in Fig 8 shows that it uses SLAM for Localization as discussed before and then also uses the Live Lidar Data to keep updating the map for real-time navigation.

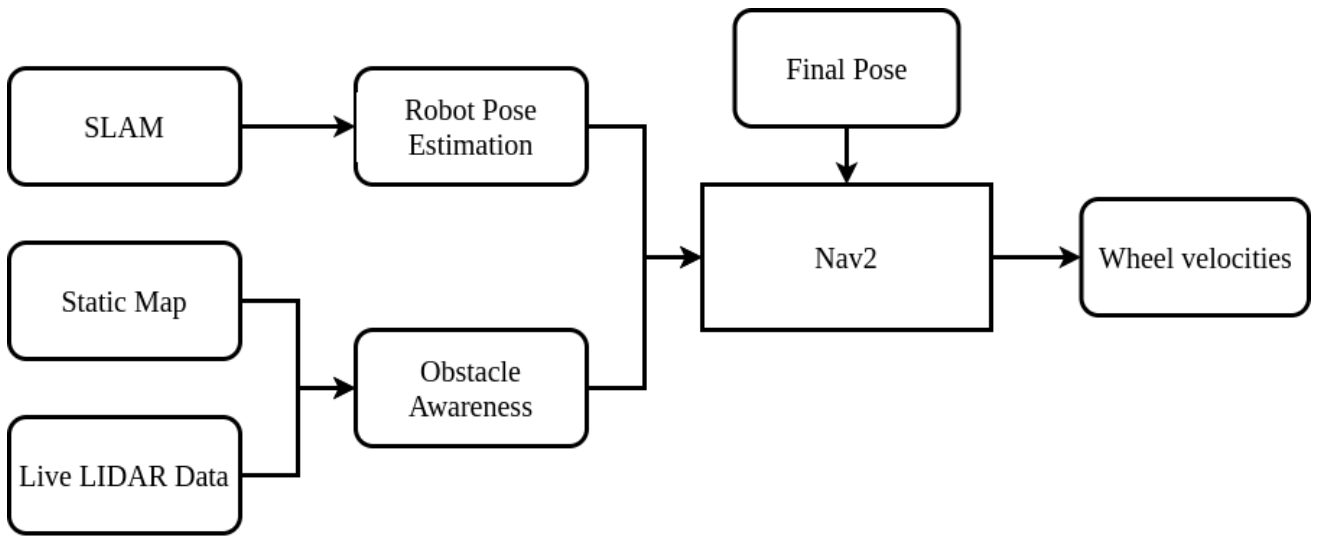

(Fig 8: Nav2 Architecture)

A standard interface has been created in the controller package so that Nav2 can work with the drone using the same interfaces exposed by it. This is done since Nav2 works with only ground robots as discussed above. Now the same Nav2 capabilities of guiding ground robots such as turtlebots can be used in drones using the standard interface of Twist Messages. Twist messages consist of 2 vectors, one denoting the linear movement in 3 axes, and the other denoting the angular movement in the 3 axes. Now all 6 degrees of freedom of the drone can be one-to-one mapped to the Twist interface. The linear movement in the x and y direction are mapped directly whereas the thrust is mapped to the linear movement in the z-direction. The roll, pitch and yaw are directly mapped to the angular movement in the x, y and z directions as can be seen in Fig 9.

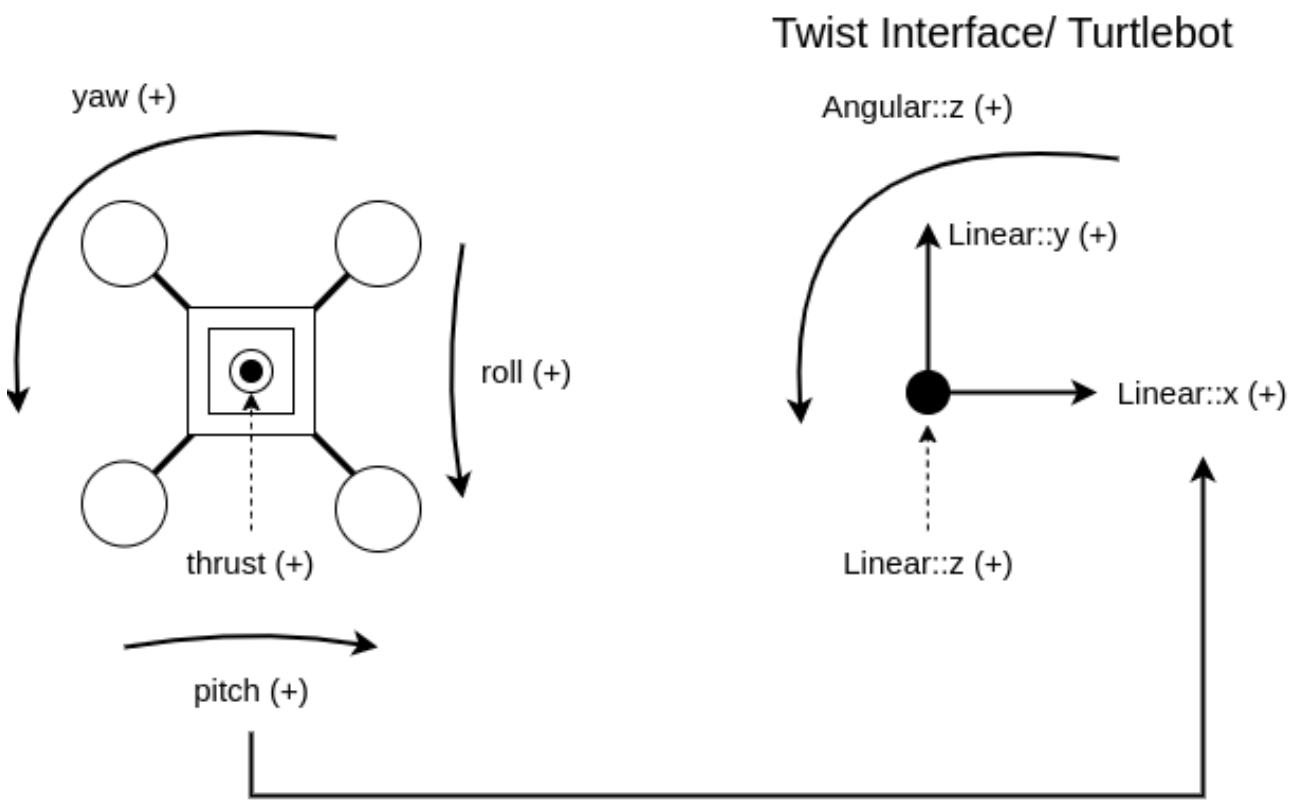

(Fig 9: Twist Interface and Mapping)

Since the drone is now hovering at a level, the linear movement in the z direction can be completely ignored. On a deeper note, it can also be seen that the drone itself can be controlled like a turtlebot in this plane. A turtlebot can be controlled by only 2 variables, one is the movement in the x direction and the angular z direction. So even in a drone, any position can be achieved using the corresponding linear x movement and the yaw movement. Bearing in mind the yaw movement is the most stable in the drone, it is the safest bet that one can manoeuvre to the required location precisely and safely. Nav2 doesn’t know that it is controlling a drone, it feels that it is controlling a turtlebot using Twist, whereas an apt mapping is made which has helped translate commands from turtlebot to drone easier. Now the actuation of the motors can be controlled by the hardware interface.

## G. Communication Structure

All the nodes and the topics in this framework can be visualized using the rqt_graph tool. It is a great tool to visualize the entire communication process and to get a general overview of it.

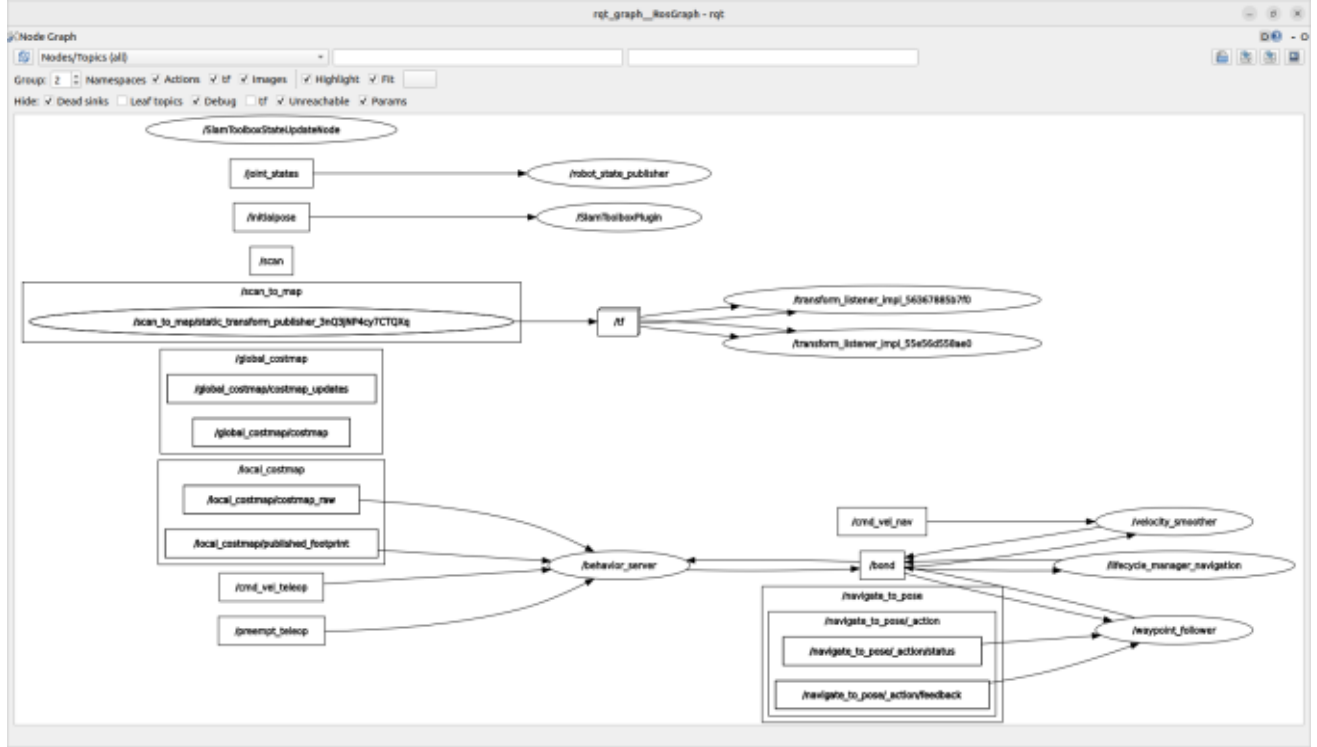

(Fig 10: rqt_graph)

A function view_frames from the tf tool can be used to visualize the transform tree which shows all the transformations discussed above which are done by the robot state publisher and the rf20_laser_odometry packages. All the communication structures have been pretty much explained in the previous sections and all of them can be visualized in the rqt_graph and Transform Tree in Fig 10 and Fig 11 respectively.

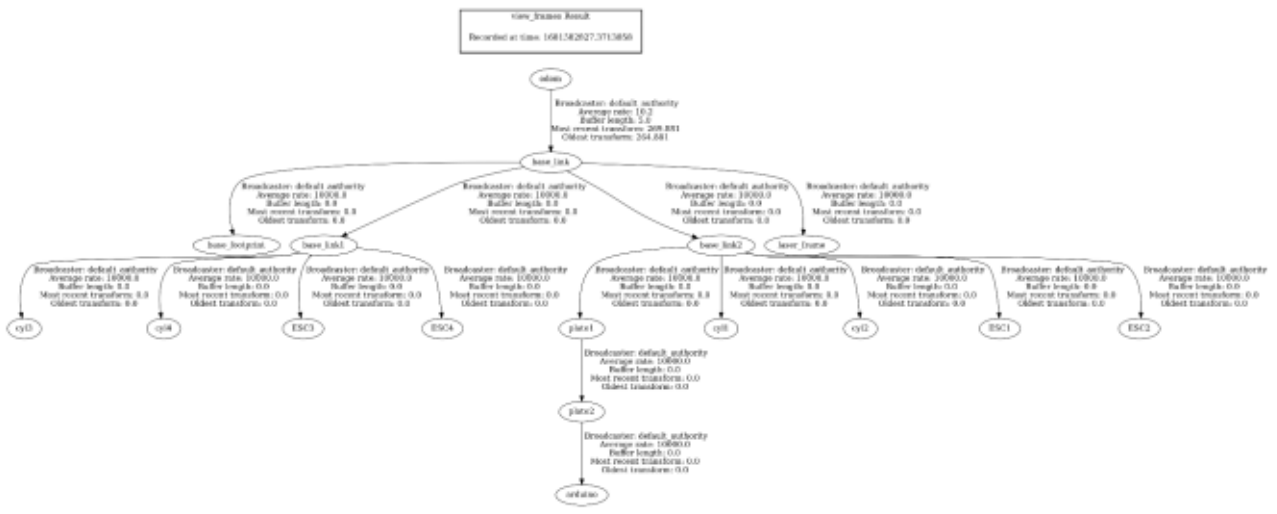

(Fig 11: Transform Tree)

## SECTION 4:

## RESULTS AND DISCUSSION:

When the grid is manually traversed using the drone, SLAM updates the map automatically. In Fig 12 at the hover level, the map generated can be seen.

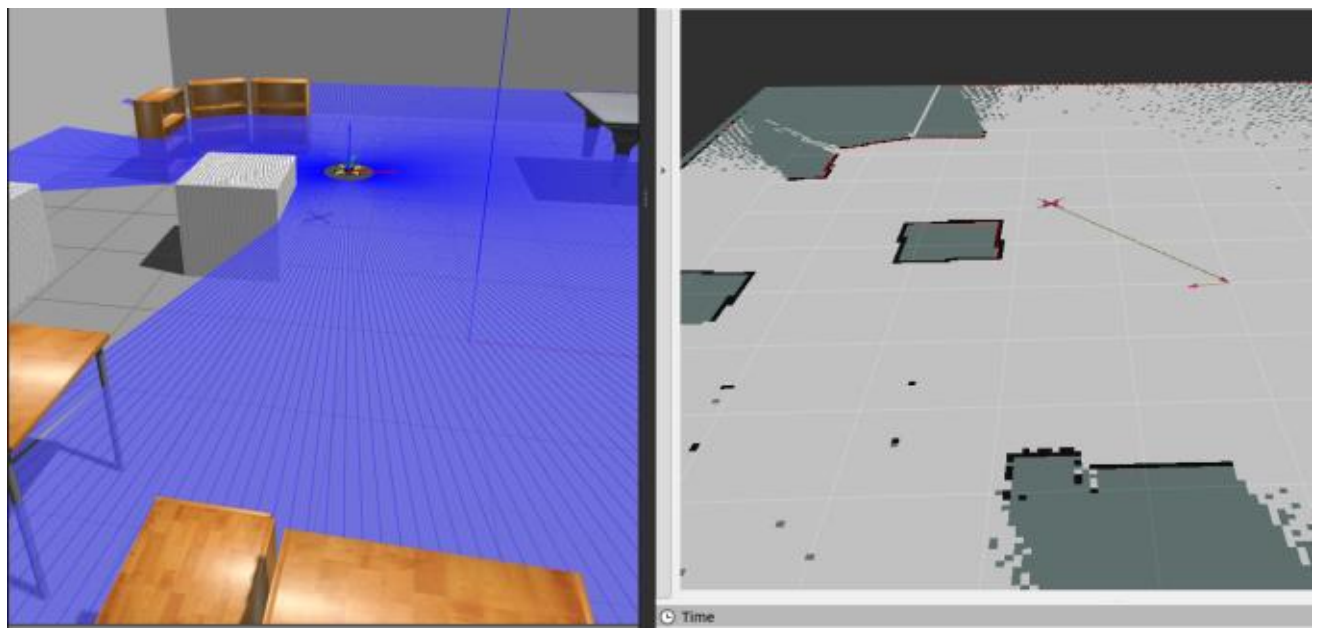

(Fig 12: 2D SLAM at Drone Hover Level)

As discussed above a corresponding cost map (Fig 13) is generated by the Nav2 package using the occupancy grid map generated by the SLAM toolbox.

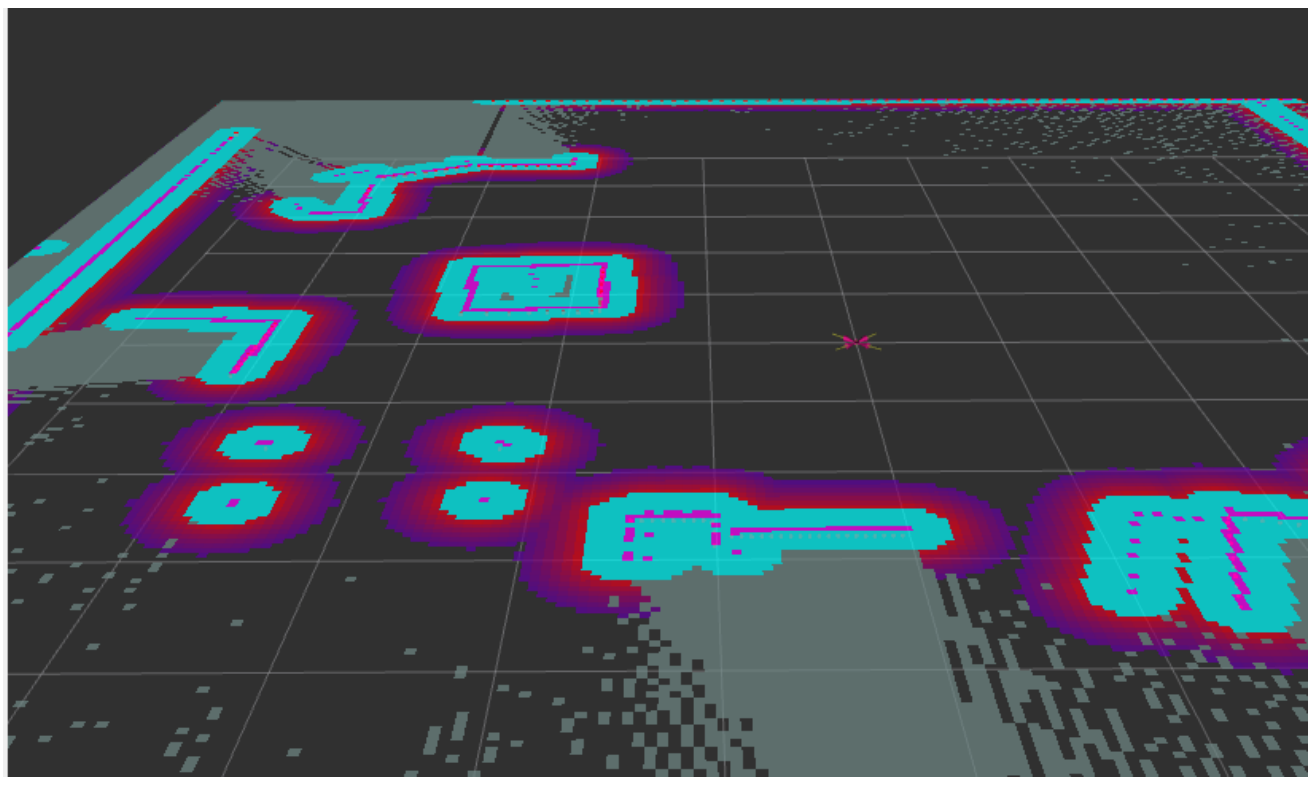

(Fig 13: Costmap)

Now that the map is saved the simulation is reset. On starting the simulation again the drone can be seen hovering at an altitude and stabilizing itself using the PID controller designed. As seen in Fig 14. Once the position is stabilised a pose estimate is provided in RViz for Nav2 to identify the drone position.

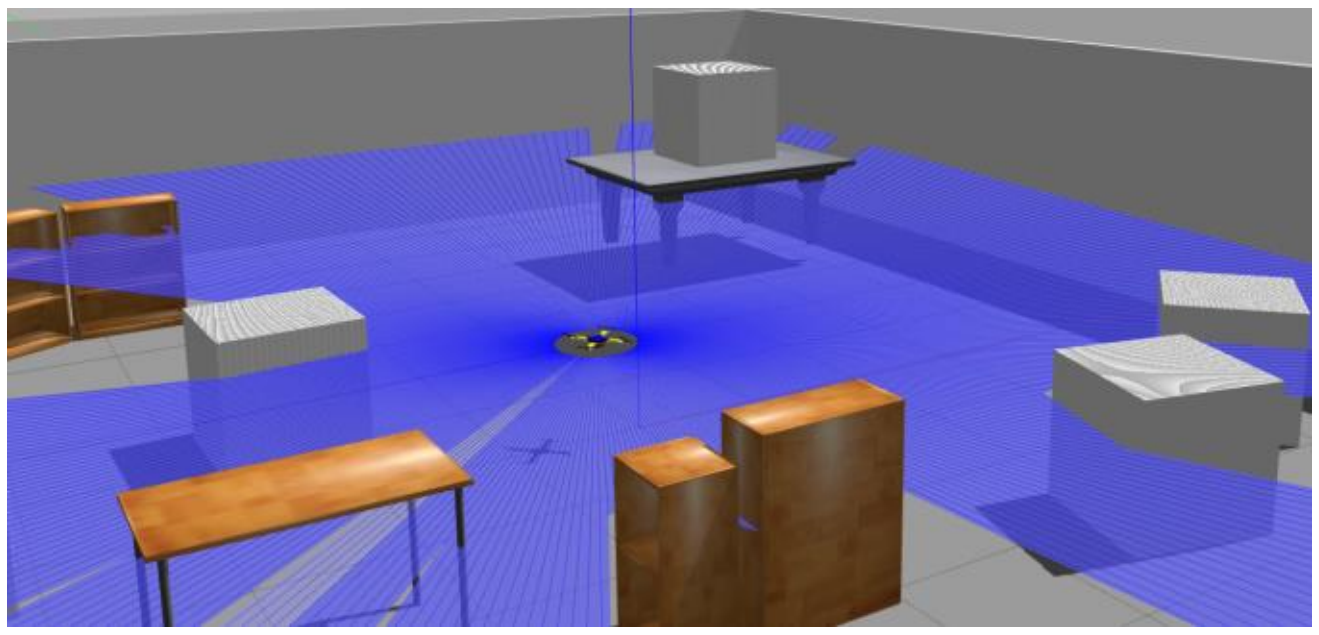

(Fig 14: Drone Initial Position)

Once the map is loaded, the destination pose has to be given to the drone using RViz. Terminal commands can also be used, but RViz makes it easier as shown in Fig 15. Green arrows indicate the location and orientation of the drone that is needed and a Pose message is sent to the Navigation System to implement.

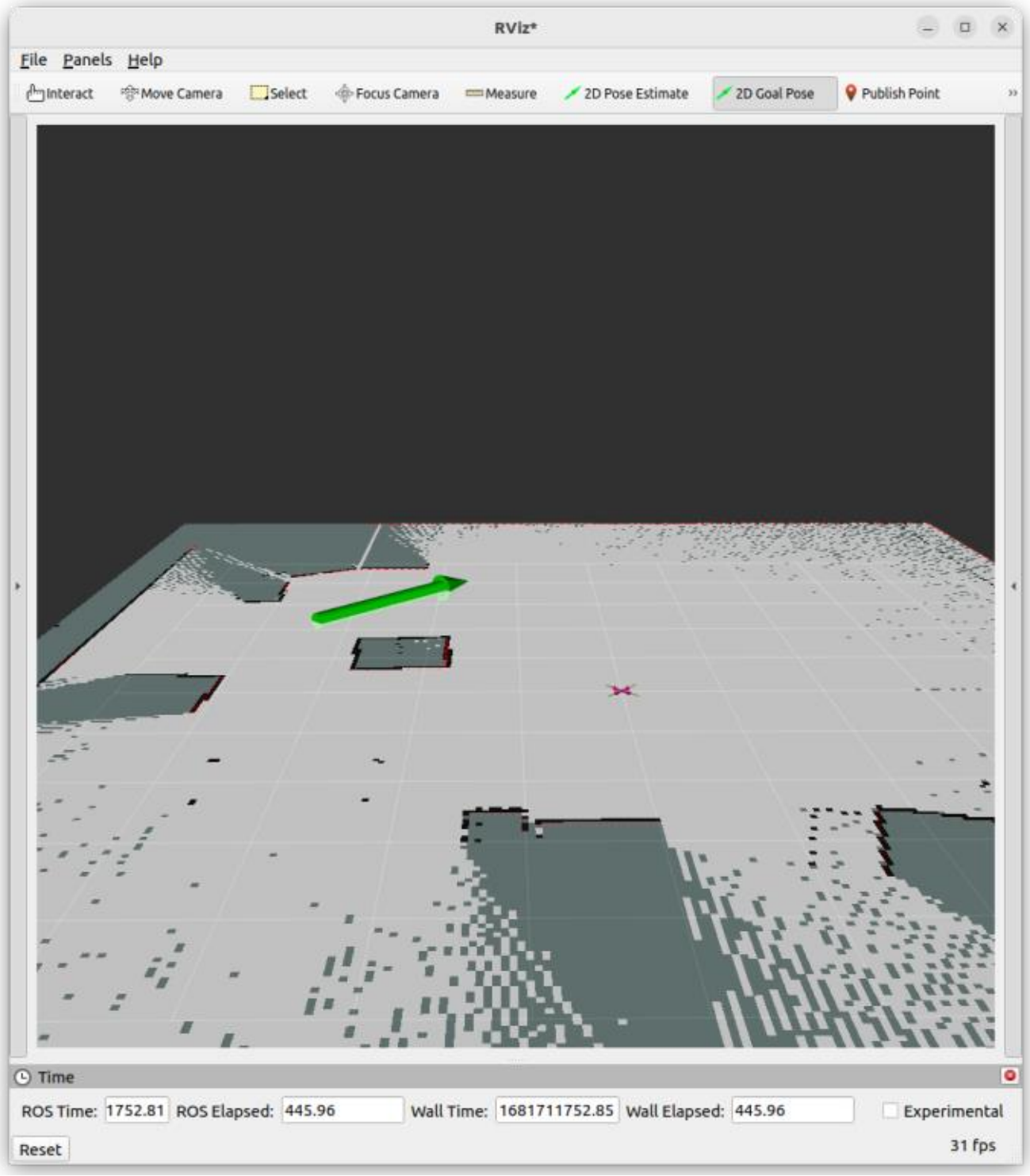

(Fig 15: Giving Destination Pose in RViz)

Once done, the drone can be seen autonomously navigating in the Gazebo environment using pitch and yaw movements to reach the destination which can be seen in Fig 16.

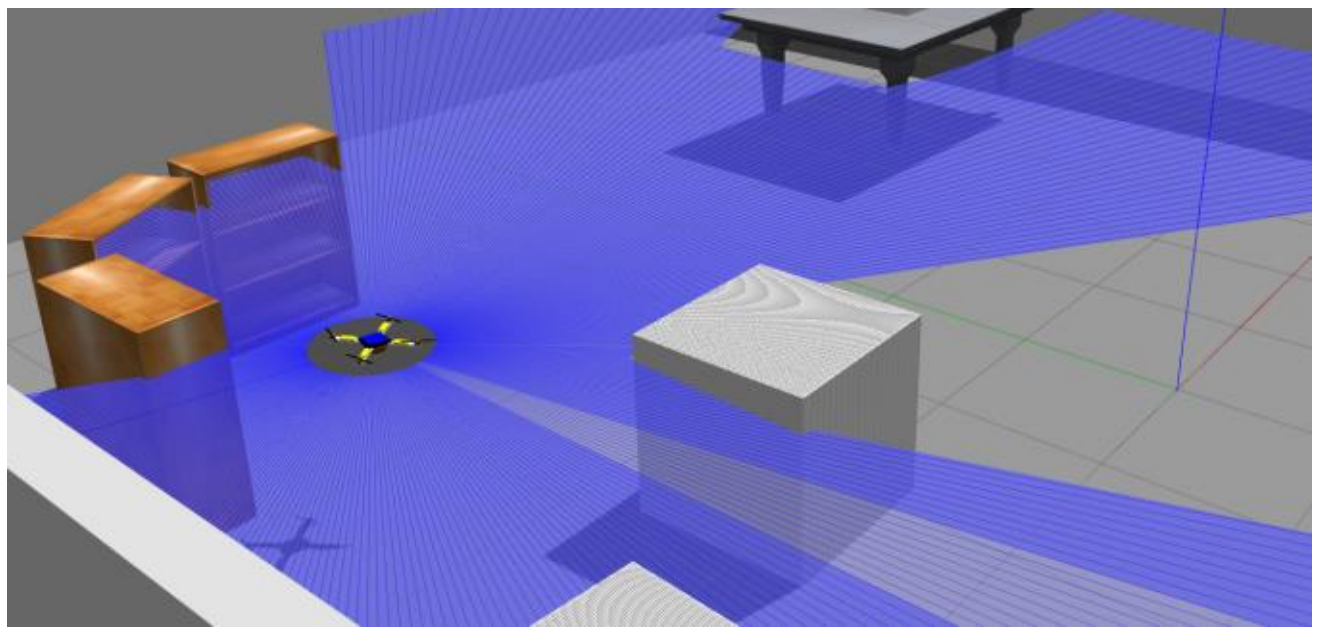

(Fig 16: Drone Reaching Destination)

## CONCLUSION:

In this study, a simulated drone built on the DJI-f450 frame was equipped with an autonomous navigation system for an Indoor GPS-free Environment and discussed the need for it. The ROS Gazebo ecosystem was used to build the framework for the project. Different ROS packages solving different problems were used and the important ones being Robot State Publisher, SLAM Toolbox, and Navigation2. With the help of a PID controller, the UAV was autonomously navigated in 2D at a fixed height using the current Navigation and SLAM systems for ground robots. Finally, the localization, mapping and autonomous navigation results were shown. The system can be enhanced with a 3D approach using 3D LiDARs and in appropriate 3D Navigation System which will provide more sophisticated manoeuvres for the quadcopter. To further gauge the effectiveness of the simulated system, evaluations of real-world implementations should be carried out.

## SECTION 5: